\documentclass[10pt,twocolumn,letterpaper]{article}

\usepackage{iccv}              

\usepackage[dvipsnames]{xcolor}
\usepackage{mdframed}
\usepackage{xspace}
\usepackage{multirow}
\usepackage{array}

\newcommand{\name}[0]{\textsc{Skald}\xspace}
\usepackage{xcolor}
\newcommand{\lu}[1]{\textcolor{black}{#1}}

\usepackage{adjustbox}
\usepackage{float}

\definecolor{cvprblue}{rgb}{0.21,0.49,0.74}
\usepackage[pagebackref,breaklinks,colorlinks,citecolor=cvprblue]{hyperref}

\def\paperID{11265} 
\def\confName{ICCV}
\def\confYear{2025}

\title{\name: Learning-Based Shot Assembly
for Coherent Multi-Shot Video Creation}

\author{
Chen-Yi Lu$^{1,2}$ \quad 
Md Mehrab Tanjim$^{2}$ \quad
Ishita Dasgupta$^{2}$ \quad
Somdeb Sarkhel$^{2}$ \\
Gang Wu$^{2}$ \quad
Saayan Mitra$^{2}$ \quad
Somali Chaterji$^{1}$ \\[2mm]
$^{1}$Purdue University \quad
$^{2}$Adobe Research
}

\begin{document}
\maketitle
\begin{abstract}
    We present \name, a multi-shot video assembly method that constructs coherent video sequences from candidate shots with minimal reliance on text. Central to our approach is the Learned Clip Assembly (LCA) score, a learning-based metric that measures temporal and semantic relationships between shots to quantify narrative coherence. We tackle the exponential complexity of combining multiple shots with an efficient beam-search algorithm guided by the LCA score. To train our model effectively with limited human annotations, we propose two tasks for the LCA encoder: Shot Coherence Learning, which uses contrastive learning to distinguish coherent and incoherent sequences, and Feature Regression, which converts these learned representations into a real-valued coherence score. We develop two variants: a base \name model that relies solely on visual coherence and \name-text, which integrates auxiliary text information when available. 
    Experiments on the VSPD and our curated MSV3C datasets show that \name achieves an improvement up to \textbf{48.6\%} in IoU and a \textbf{43\%} speedup over the state-of-the-art methods. A user study further validates our approach, with \textbf{45\%} of participants favoring \name-assembled videos, compared to 22\% preferring text-based assembly methods.
    Project URL: \url{https://schaterji.io/publications/2025/SKALD}
\end{abstract}
\section{Introduction}
\begin{figure}[t]
    \centering
    \includegraphics[width=1\linewidth]{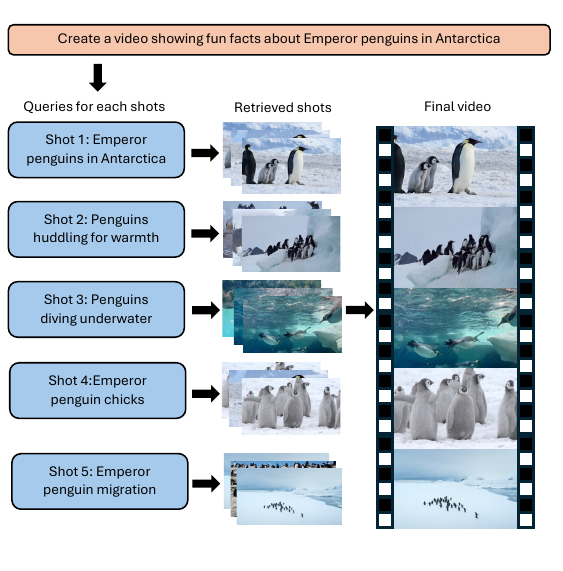}
    \caption{\textit{Illustration of a multi-shot video assembly pipeline}. 
     Each shot is defined by a query (e.g., ``Penguins huddling for warmth''), used to retrieve candidate clips from a large video pool. The final sequence is assembled by selecting clips that best match each query while ensuring visual and thematic coherence, resulting in a concise video on Emperor penguins in Antarctica.}
    \label{fig:assembly_problem}
\end{figure}
\begin{figure}[t]
    \centering
    \includegraphics[width=\linewidth]{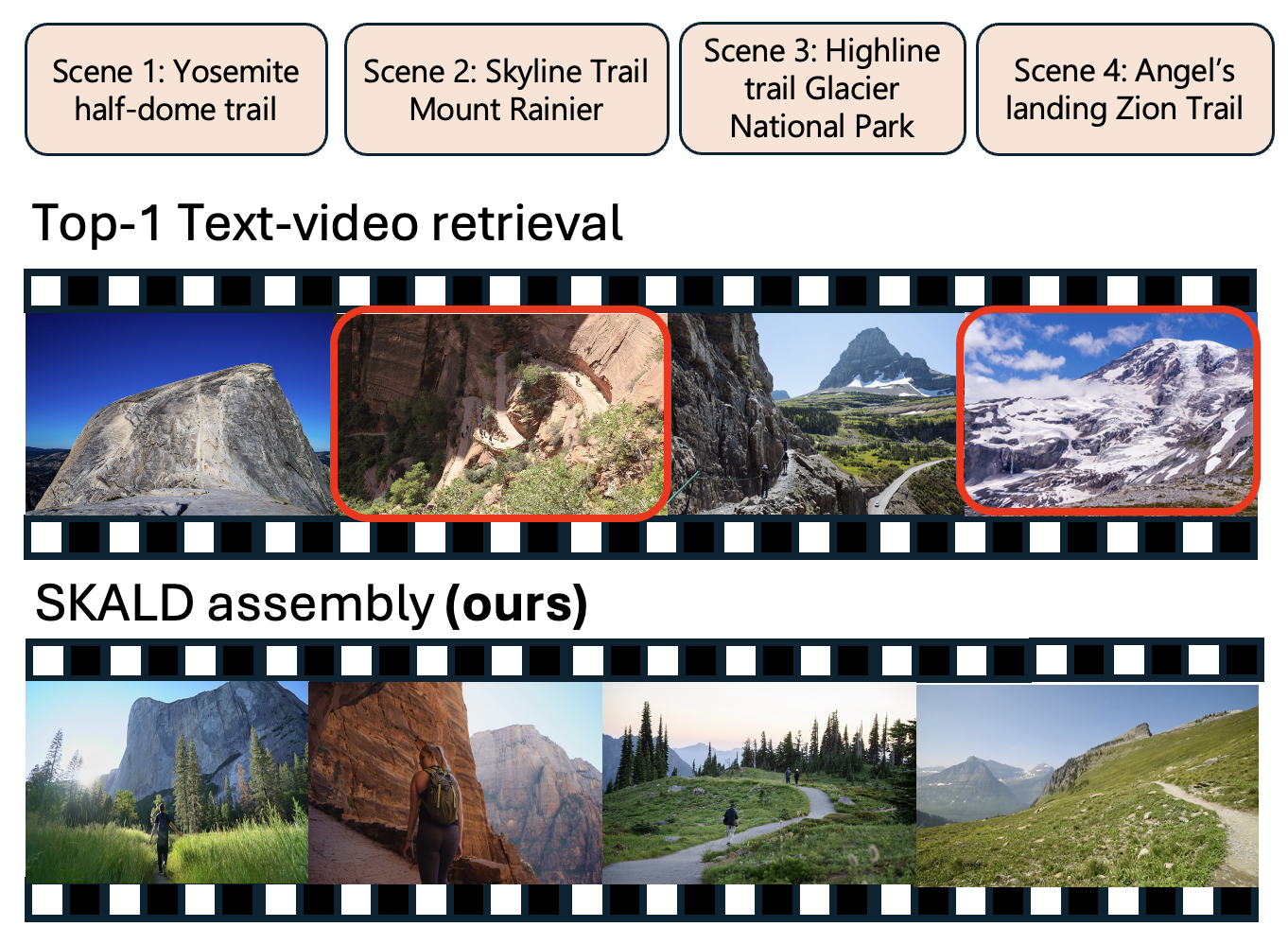}
    \caption{
    \textit{Challenges in multi-shot video assembly}. Text-based retrieval 
    selects relevant shots independently but may create incoherent sequences
    (\textit{e.g.}, a snow scene amid summer footage).
    }
    \label{fig:compare}
\end{figure}

Short-form video platforms have sparked a surge in user-generated content, with countless individuals creating and sharing edited videos. However, assembling a high-quality multi-shot video from raw or stock footage remains challenging. 
Capturing aesthetically pleasing scenes often demands advanced skills, equipment, and the right locations, and even with an abundance of stock clips available, the editing process requires significant expertise and effort ~\cite{arijon1976grammar, musser2012early}.
Amateur users frequently find multi-shot video assembly daunting, and professionals must invest extensive time to sift through footage for a coherent narrative. 
This gap motivates our research question: 
\textit{Can an artificial system assist in assembling a coherent multi-shot video based on a learned notion of narrative coherence?}

We formally define the multi-shot video assembly problem as follows: suppose we want to create a video containing $n$ shots, each shots has $k$ candidates. The task is to assemble the most coherent sequence of shots. Optionally, each shot may be accompanied with text descriptions or visual keywords. As illustrated in~\cref{fig:assembly_problem}, the goal is to ensure logical flow and visual appeal in assembled video. 
%
A common strategy is to cast multi-shot assembly as a text-to-video retrieval problem~\cite{jiang2022tencent,miech2020end,bain2021frozen,lin2023video,luo2021clip4clip,deng2023prompt, wu2023cap4video}, selecting each shot by matching text descriptions. While this method can work in scenarios where textual information is available, it presents significant challenges. 
First, as shown in~\cref{fig:compare}, individual matched shots can be semantically inconsistent. 
Previous works~\cite{truong2016quickcut,wang2020story,wang2019write,xiong2022transcript,yang2023shot,lu2023show} attempted to address this issue using shot-level captions and by training either global aggregation networks or cross-attention~\cite{lin2022cat} mechanisms to assemble sequences based on these captions. However, these methods are constrained by the limited availability of video-caption pairs and are often rigid, being tied to specific datasets.  
Furthermore, these approaches struggle in scenarios where captions are unavailable. 

Instead, we propose a more flexible, ranking-based approach that explicitly evaluates combinations of shots using a dedicated \emph{coherence} metric. Although substantial research exists in no-reference video quality assessment (VQA)~\cite{xing2022starvqa,xing2023starvqa+,kou2023stablevqa,wu2022fast,wu2023exploring,wu2023q}, these methods primarily focus on attributes such as resolution or aesthetic quality, often overlooking semantic flow between consecutive shots. Consequently, high-quality but logically disconnected footage might be rated favorably, while sequences with slightly lower visual quality but stronger narrative coherence could be unjustly penalized. This gap motivates the need for a specialized method that explicitly quantifies shot-level continuity and thematic consistency.
To address these challenges, we propose \name, a multi-shot video assembly method comprising two key components:
\noindent (1) the \emph{Learned Clip Assembly (LCA)} score, a learning-based assessment metric specifically designed to evaluate multi-shot coherence, modeling both temporal and spatial relationships among shots, and
\noindent (2) an efficient assembly algorithm that leverages the LCA score to mitigate the exponential complexity of assembling coherent video sequences.
We train a transformer-based encoder on the visual embeddings of each multi-shot sequence to capture temporal and spatial relationships across shots. A regression layer then transforms the resulting feature vector into a real-valued coherence measure, the LCA score, which guides the selection of the most coherent shot arrangement.

\lu{
To train our LCA encoder with minimal annotations, we introduce a contrastive \emph{shot coherence learning} task. 
Positive pairs are constructed from shots in the same reference video, with minor frame-level augmentations that do not disrupt visual continuity, while negative pairs are generated by replacing certain shots with visually disjoint content to break continuity. After training the LCA encoder on this task, we freeze its weights and train a small regression head on a limited set of human ratings. Our findings show that this two-stage approach (contrastive pretraining plus light supervision) significantly outperforms end-to-end models, even under limited annotation budgets.}
\lu{Exhaustively searching through all shot combinations using the LCA score is infeasible ($\mathcal{O}(k^n)$ complexity). We regularize the model to ensure the learned LCA score remains invariant to sequence length, enabling a \emph{beam search} approach that efficiently locates high-scoring shot combinations without exploring every possibility.
In addition to this visual-only pipeline, we also introduce \name-text, which leverages textual cues (\textit{e.g.}, caption-shot similarity) when available, in tandem with the LCA score, to further improve assembly quality. 
We further curate a multi-shot video dataset, \emph{MSV3C}, from the V3C dataset~\cite{rossetto2019v3c} for evaluating \name\ alongside existing text-based assembly methods~\cite{xiong2022transcript, yang2023shot} and video foundation models~\cite{luo2021clip4clip, bain2021frozen, radford2021learning}.}
In short, our contributions 
are as follows:
\begin{enumerate}[leftmargin=1.5em]
    \item \textbf{\name}, a multi-shot video assembly method powered by the \textbf{LCA score}, a novel learning-based metric capturing temporal and spatial shot relationships.
    \item A two-stage training scheme:
    \textbf{(i)} \emph{Shot Coherence Learning} (contrastive) to distinguish coherent vs.~incoherent sequences, and
    \textbf{(ii)} \emph{Feature Regression} to 
    derive coherence scores from limited human annotations.
    \item A beam search algorithm with sequence-length decorrelation that makes multi-shot assumbly more scalable and efficient than exhaustive search.
    \item \textbf{\name-text}, a variant incorporating textual cues when available to boost assembly quality.
    \item Extensive evaluations on the VSPD~\cite{yang2023shot} and our MSV3C datasets demonstrate improvements of up to \textbf{48.6\% in IoU} and \textbf{43\% faster inference} compared to state-of-the-art methods~\cite{yang2023shot, luo2021clip4clip, xiong2022transcript}, with a user study confirming a \textbf{45\% participant preference} for our approach versus 22\% for text-based methods~\cite{luo2021clip4clip}.
    \end{enumerate}


\section{Related Work}

\subsection{Text-guided Shot Assembly}
Several works have addressed the problem of shot assembly through text-to-video retrieval~\cite{yang2023shot, wang2019write, lu2023show, xiong2022transcript}. 
Given text descriptions, either at the shot level or as a full video synopsis, these approaches encode both text and video modalities into a shared representation space, finding the closest matching embeddings. 
Unlike the traditional text-video retrieval problem, this line of work also emphasizes intra-shot cohesion, ensuring that the retrieved shots are coherent with each other. For instance, Yang \etal~\cite{yang2023shot} propose transformer-based encoders that take both text and video tokens as input, training the encoder to distinguish reference videos from shuffled or randomly inserted shot sequences using cross-entropy loss. Xiong \etal~\cite{xiong2022transcript} introduce an adaptive encoder module that improves video-to-video coherence by updating text embeddings based on 
visual cues from preceding frames, enhancing coherence.
Lu \etal~\cite{lu2023show} 
further improve consistency by
aligning text and video representations at both local (shot-to-text) and global (story-level) scales. 
Despite their effectiveness, these approaches rely on expensive shot-level text annotations and custom datasets, limiting adaptability to new video domains and constraining real-world applicability.
In contrast, 
we directly score multi-shot video sequences for coherence, guiding assembly without detailed text labels. 

\subsection{Video Quality Assessment (VQA)}
Extensive work has been done on developing quality metrics for videos to assess their technical quality and aesthetics. Most approaches involve training neural networks to align with human preferences. Li \textit{et al}.~\cite{li2019quality} proposed VSFA, which incorporates content-aware features from pre-trained CNNs and 
GRU-based temporal modeling, along with
a subjectively-inspired temporal pooling layer human-centric quality perception. 
Wu~\etal~\cite{wu2022fast} developed FAST-VQA, an efficient end-to-end VQA method that uses fragment sampling to reduce computational costs while maintaining accuracy. In a more holistic approach, Wu \textit{et al}.~\cite{wu2023exploring} explored evaluating both technical and aesthetic aspects of video quality. They created the DIVIDE-3k dataset---offering diverse human opinions on video quality---and 
proposed DOVER, a model addressing both technical and aesthetic facets. Building on this, Wu \textit{et al}.~\cite{wu2023q} introduced Q-Align, leveraging large multi-modal models for visual quality assessment, demonstrating robust
generalization to various visual tasks.
However, existing VQA methods 
typically demand large-scale human preference data and assess individual shot quality rather than multi-shot coherence. As a result, well-shot yet thematically disjoint clips can receive high VQA scores despite lacking narrative structure.
To address this, we propose the Learned Clip Assembly (LCA) score, a novel metric for evaluating coherence, logical flow, and overall assembly quality. Our two-step training---contrastive pre-training followed by minimum human supervision---significantly reduces annotation needs while effectively capturing multi-shot coherence.


\begin{figure*}[ht]
    \centering
    \includegraphics[width=0.9\linewidth]{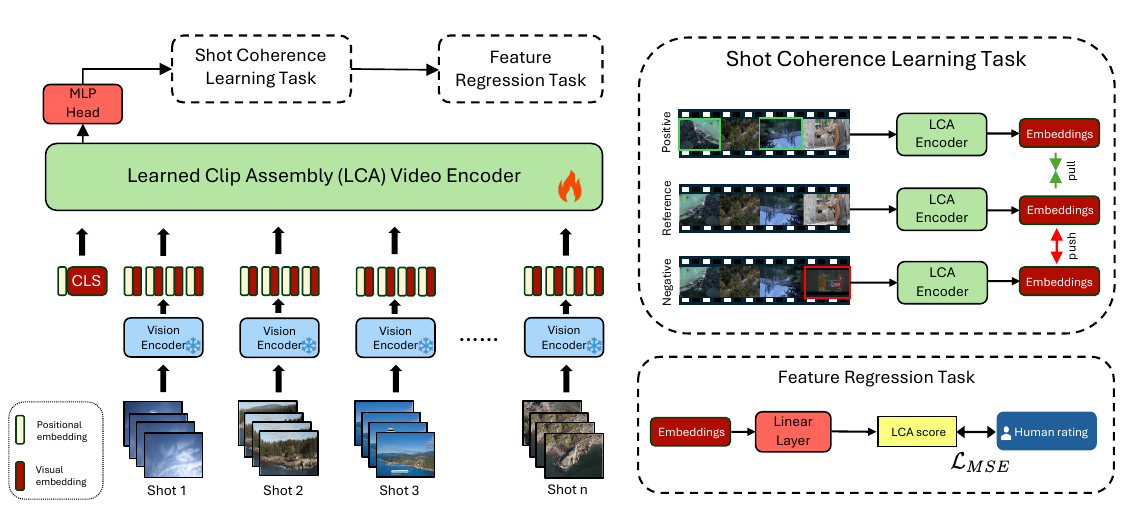}
    \caption{\textit{Training pipeline of our LCA encoder}. The LCA encoder evaluates shot coherence in multi-shot video sequences. Frames from each shot are processed by a frozen vision encoder to generate embeddings, which are combined with positional embeddings and a \texttt{CLS} token and fed into a Transformer-based LCA encoder. The final shot sequence representation is aggregated into a \(\mathbb{R}^{768}\) vector via an MLP head. Training includes the Shot Coherence Learning Task (\textit{top right}), where contrastive learning distinguishes coherent (positive) from incoherent (negative) assemblies using Noise Contrastive Estimation (NCE) loss. Positive samples are perturbed to maintain coherence, while hard-negative mining generates incoherent samples. The Feature Regression Task (\textit{bottom right}) uses a linear layer to map frozen LCA encoder features to an LCA score aligned with human ratings, trained with MSE loss. Together, these tasks equip the LCA encoder to assess the quality and narrative consistency of shot assemblies.}
    \label{fig:method-main}
\end{figure*}

\section{Approach}
\subsection{Problem Definition}
\label{subsec:dataprep}
We address \emph{multi-shot video assembly}---the task of creating coherent short videos. Formally, let us denote each short video by $V = \{s_1, s_2, \ldots, s_n\}$, where each $s_i$ represents a shot (\textit{i.e.}, a continuous take from a single camera). \emph{Impactful} promotional or social media videos 
require well-assembled shots to engage viewers and deliver a clear narrative.
However, manually selecting and arranging these shots is time-consuming, motivating automated solutions.

\noindent\textbf{Problem Setup.} Suppose we need to assemble $n$ shots into a final video, and for each shot position $i \in \{1, 2, \ldots, n\}$, we have $k$ candidate shots. The total number of possible sequences is then $k^n$, making exhaustive search computationally infeasible for large $n$. Our goal is thus to learn a function 
\(
f : V \rightarrow \mathbb{R}
\)
that assigns a \emph{coherence score} to any multi-shot sequence $V$. During inference, we use $f$ to efficiently select the most coherent arrangement of $n$ shots from the $k^n$ possibilities, without exhaustive enumeration.

\subsection{Data Preparation}
\label{sec:dataprep}
\noindent\textbf{Learning from Edited Videos.}
Prior methods often learn shot assembly via text-to-video retrieval using detailed shot-level captions, \emph{i.e.}, $\{c_i\}_{i=1}^n$ aligned with $\{s_i\}_{i=1}^n$, to supervise the model~\cite{yang2023shot,xiong2022transcript,lu2023show,wang2019write}. However, collecting such fine-grained captions is expensive and limits the model’s ability to generalize to unlabeled scenarios. Instead, we directly learn from \emph{already edited} videos, which we assume to be coherent by construction. Given each professionally curated video $V$, we treat the entire sequence as a positive example of coherent assembly. We then train $f$ to rate these sequences in a manner consistent with human judgments of quality and coherence.

\begin{figure}
    \centering
    \includegraphics[width=1\linewidth]{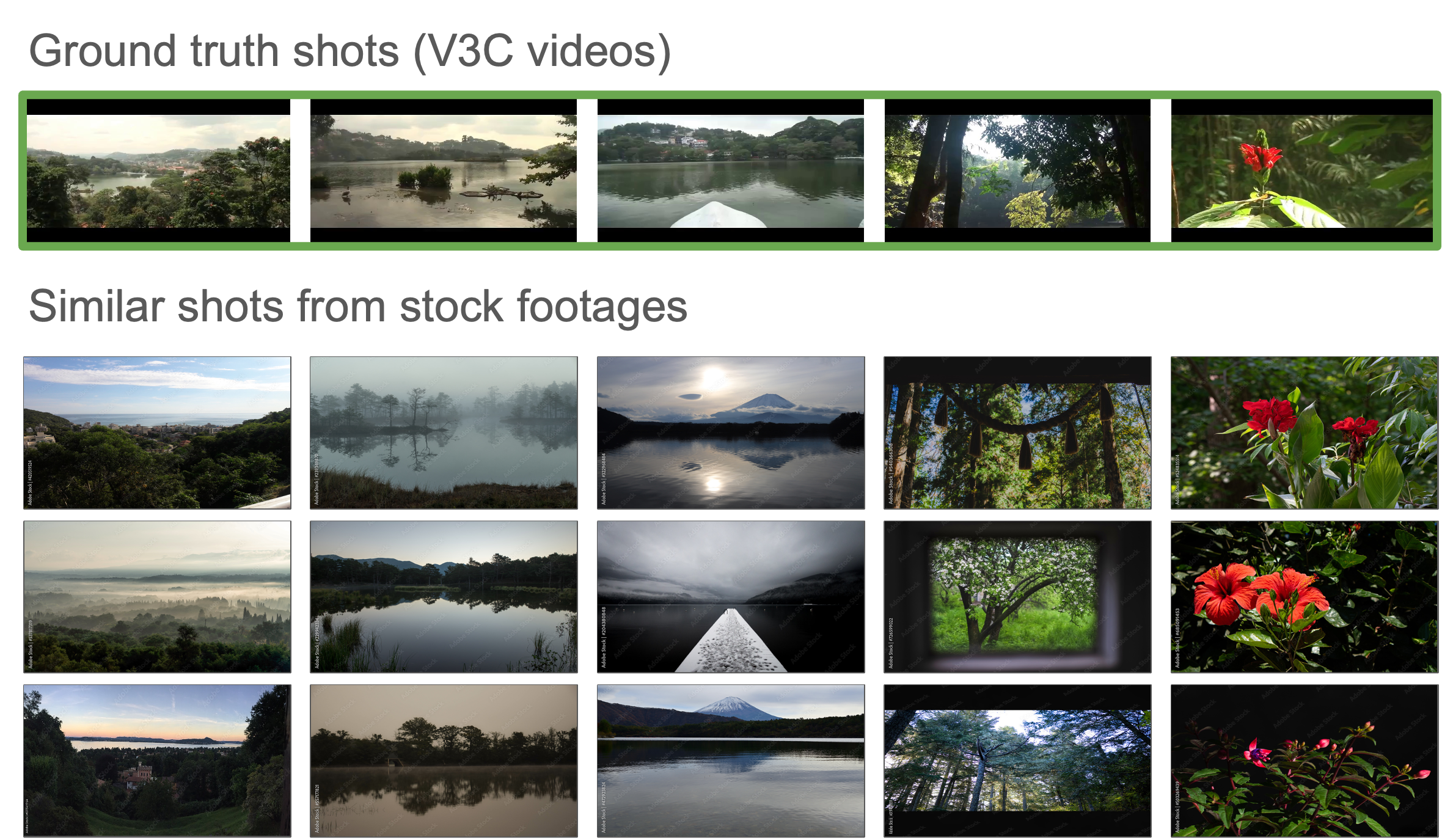}
    \caption{\emph{MSV3C testing set construction.} To simulate real-life video editing, we generate multiple candidate shots for each
    ground-truth shot. We first generate captions for the ground-truth shots and then use these captions to retrieve visually similar alternatives from a stock footage collection.}
    \label{fig:MSC3Ctest}
\end{figure}

\noindent\textbf{Constructing the \textit{MSV3C} Dataset.}
To train and evaluate our method, we begin with the V3C dataset~\cite{rossetto2019v3c}, a large-scale repository of over $28,450$ videos. After excluding formats unsuitable for multi-shot assembly (e.g., talking-heads, single-take vlogs, or music videos), we segment the remaining $5,648$ videos into individual shots using TransNetv2~\cite{souvcek2020transnet}, ultimately yielding over $12,000$ shots. We then remove shots under two seconds, excessive text overlays, and those with minimal frame variance (as measured by CLIP-based embedding diversity), leaving a refined pool of videos we refer to as \textbf{MSV3C}. From this pool, we select $500$ videos with balanced shot counts for detailed annotation. Each video is rated by human annotators on a 5-point scale measuring \emph{shot coherence} (how smoothly shots flow) and \emph{semantic consistency}, with $1$ indicating highly disjointed sequences and $5$ indicating seamless transitions. Each annotator evaluates at least $20$ videos (including three “golden” references), and we discard ratings from any annotator who scores below 4 on a golden video. We further remove outlier ratings, then average the remaining valid annotations to form a ground-truth coherence score for each video. The supplementary material provides additional details on this data-creation and filtering process.

For evaluation, we split these curated MSV3C videos on a $9:1$ ratio and annotate each shot with a concise scene description. We then retrieve five visually and semantically similar shots (using CLIP-based queries) from our stock footage pool for each original shot. Our evaluation objective is to see how well each method reconstructs the original, most coherent shot sequence. As depicted in \cref{fig:MSC3Ctest}, although each retrieved shot is semantically aligned with its target, the original shot arrangement typically achieves superior visual flow.

\vspace{-0.1in}
\subsection{Learning the LCA Score}
\label{subsec:lca_score}

At the heart of our approach is the \emph{Learned Clip Assembly (LCA) score}, a novel metric designed specifically to assess how well a multi-shot sequence flows and maintains narrative coherence. In contrast to video quality metrics focusing on purely aesthetic or technical attributes, our LCA score directly captures the sense of smooth transitions and logical sequencing among shots. The overview of the training of LCA encoder is shown in~\cref{fig:method-main}. 
Each shot is independently processed by a frozen vision encoder to obtain frame-level embeddings, which are then aggregated—together with a learnable \texttt{[CLS]} token—into shot-level representations. Positional embeddings are appended to preserve temporal information. These shot embeddings are fed into the LCA video encoder, which refines the temporal and semantic relationships among shots to produce a cohesive sequence representation.
On top of the LCA encoder, we attach a lightweight MLP head for (1) the \emph{shot coherence learning} task, which uses contrastive learning with positive and negative samples, and (2) a \emph{feature regression} task, which aligns the learned representations with human-annotated coherence scores.

\noindent\textbf{Shot Coherence Learning Task.}
To preserve storytelling in assembled videos, we treat the problem of maintaining coherent shot transitions as a self-supervised contrastive learning task. Given a coherent, professionally edited video $V = \{s_1, s_2, \dots, s_n\}$ that already exhibits smooth narrative flow, we generate two types of samples: \emph{positive} and \emph{negative}. The positive sample $V^+$ is derived by applying light, frame-level perturbations (e.g., small random crops or brightness shifts) that do not compromise scene continuity. These adjustments primarily affect low-level pixel information without altering the overall composition or thematic connections among shots. As a result, the perceived visual flow remains intact, and the storytelling aspect of $V^+$ is preserved. 

On the other hand, the negative sample $V^-$ is constructed by replacing a subset of shots in $V$ with content that is visually and contextually disjoint, determined by a low CLIP-based similarity score. These abrupt “inserts” serve to break the story flow, introducing thematically inconsistent transitions (e.g., jumping from an indoor cooking shot to a completely unrelated outdoor action scene). Our model thus learns to distinguish seamless transitions from jarring cuts, even when the raw shot content is more heterogeneous than in a typical cinematic production. By focusing on continuity at a higher, scene-to-scene level—rather than subtle, frame-by-frame cues—the method is robust to variations that might occur in real-world videos, such as changes in camera angle or shot duration.

Formally, we optimize via noise contrastive estimation (NCE), minimizing
\begin{equation}
  \small
  \mathcal{L}_\text{NCE} = 
   -\log 
   \frac{\exp\!\bigl(\tfrac{s(V, V^+)}{\tau}\bigr)}
        {\exp\!\bigl(\tfrac{s(V, V^+)}{\tau}\bigr) + \sum_{V^- \in \mathcal{N}} 
         \exp\!\bigl(\tfrac{s(V, V^-)}{\tau}\bigr)},
    \label{eq:NCE}
\end{equation}
where \(s(\cdot,\cdot)\) denotes a cosine similarity measure in the learned embedding space, \(\tau\) is a temperature hyperparameter controlling the concentration of the distribution, and \(\mathcal{N}\) is the set of negative samples. By pulling together $V$ and $V^+$—which differ at the pixel level but preserve narrative structure—our model emphasizes features that contribute to a coherent story. Simultaneously, pushing apart $V$ and $V^-$ trains the system to discount sequences with abrupt thematic shifts. Crucially, this does not require shot-level text labels; the only human-annotated “label” is a coherent original video $V$, enabling large-scale data collection. In this manner, the model learns a robust notion of narrative flow, ensuring that shot assemblies align with the underlying storyline and avoid disjoint content that disrupts viewer engagement.

\noindent\textbf{Feature Regression Task.}
After training the LCA encoder on the shot-coherence task, we freeze its weights and train a lightweight regressor to map the encoder’s output to a real-valued score aligned with human ratings (Section~\ref{subsec:dataprep}). Concretely, we pass the final sequence-level embedding through a linear layer to produce the predicted score \(\hat{y}_i\) for the \(i\)-th sample, with \(y_i\) as the ground-truth rating. We then optimize using the mean squared error (MSE) loss, where N is the number of samples:
\begin{equation}
\label{eq:mse}
    \mathcal{L}_{\text{MSE}} 
= \frac{1}{N} \sum_{i=1}^N \bigl(\hat{y}_i - y_i\bigr)^2,
\end{equation}
\noindent\textbf{Two-Stage Training vs. End-to-End.}
A key design choice in our framework is the sequential training procedure: first optimizing \(\mathcal{L}_\text{NCE}\) to learn multi-shot coherence representations at scale, then freezing the encoder and fine-tuning a lightweight regressor with \(\mathcal{L}_{\text{MSE}}\). This two-stage approach contrasts with an end-to-end scheme that jointly optimizes both losses (\ie training with $\mathcal{L}_\text{NCE} + \mathcal{L}_\text{MSE}$). Training \(\mathcal{L}_\text{NCE}\) alone requires only unedited (coherent) videos—no extensive shot-level annotations—thus allowing us to leverage large-scale, minimally labeled data for representation learning. Only in the second stage do we need a moderate set of human-rated coherence scores to align the learned embeddings with actual perception of video flow. By isolating the regression task from feature learning, our method not only reduces the required volume of labeled data but also stabilizes the optimization process, avoiding the data-intensive demands of a fully end-to-end approach.

\begin{figure}
    \centering
    \includegraphics[width=1\linewidth]{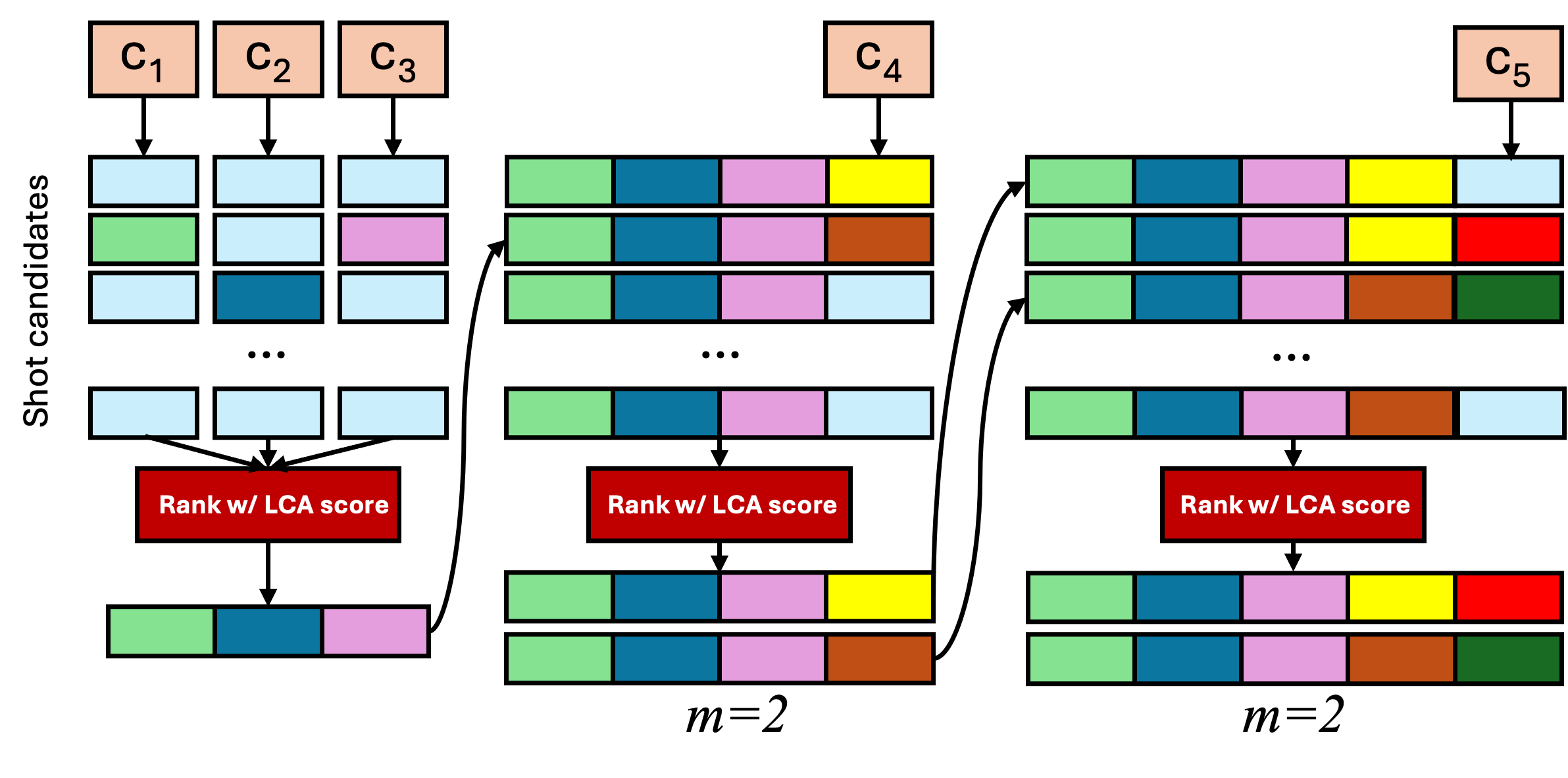}
    \caption{Beam search in \name. 
    At each step, \name evaluates partial shot sequences using the LCA score, retaining only the top-$m$ ranked sequences for extension. This iterative process ensures coherent shot selection without exhaustive search. 
    We demonstrate \name's beam search pipeline with \(m=2\).}
    \label{fig:beam}
\end{figure}

\subsection{\name: Multi-Shot Video Assembly}
\label{sec:inference}

In Section~\ref{subsec:dataprep}, we established that the time complexity of brute-force search across all possible shot combinations is $\mathcal{O}(k^n)$, where $k$ is the number of candidate shots per position and $n$ is the total number of shots in the target video. As this complexity is prohibitively expensive for practical applications, we propose a beam search approach to efficiently explore the solution space~\cite{lowerre1976harpy, freitag2017beam}.

\noindent\textbf{\name.}
Our algorithm, \name, operates as shown in \cref{fig:beam}. Given the task of creating an $n$-shot video from $k$ candidate shots per position—optionally captioned by \(\{c_i\}_{i=1}^n\)—we first evaluate the LCA scores for all possible triplets of the first three shots, $\{s^1_i, s^2_j, s^3_l\}$ with $i,j,l \in \{1,\ldots,k\}$. We maintain a beam of size $m$, storing only the top-$m$ sequences by LCA score. For each subsequent position $t$ ($4 \le t \le n$), we extend every sequence in the beam with each candidate shot $s^t_j$ ($1 \le j \le k$), compute their LCA scores, and then retain the top-$m$ scoring sequences. By adjusting $m$, we can balance between video quality and computational cost.

Since the LCA score should be valid for sequences of varying lengths, we introduce a \emph{sequence length decorrelation} term during training. Specifically, for each batch we compute the covariance between the normalized LCA output embeddings and the normalized sequence lengths, and penalize any correlation~\cite{kim2019learning,bahng2020learning,lee2021learning}. This discourages the model from associating specific score ranges with specific lengths, leading to more consistent scoring across short or long sequences.

\noindent\textbf{\name-text.}
While \name relies solely on the visual LCA score, some real-world editing scenarios also include textual inputs (e.g., user-provided scripts or shot-level descriptions). To accommodate this, we propose a variant called \name-Text. In addition to the LCA score, we compute a cross-modal similarity for each candidate shot $s^t_j$ and its corresponding text caption $c^t_j$ (if available) using a pretrained text--image model (e.g., CLIP). Let 
\(
\mathrm{Sim}_\text{clip}\bigl(s^t_j,\,c^t_j\bigr)
\)
denote this text-based similarity. We then combine it with the LCA score in the beam search:
\[
S_{\mathrm{combo}}\bigl(V \oplus s^t_j\bigr)
\;=\;
\mathrm{LCA}\bigl(V \oplus s^t_j\bigr)
\;+\;
\gamma \,\cdot\, \mathrm{Sim}_\text{clip}\bigl(s^t_j,\,c^t_j\bigr),
\]
where $V \oplus s^t_j$ denotes the extended partial sequence. A scalar hyperparameter $\gamma$ controls the relative importance of text alignment. Unlike the beam-search inference in~\cite{xiong2022transcript}, \name and \name-text leverage CLIP-based embeddings instead of a domain-specific encoder fine-tuned for a particular dataset, enabling broader generalization. Furthermore, our design choices—precomputed embeddings, flexible textual integration, and broad-domain embeddings—enhance the scalability and applicability of our beam-search approach compared to prior work.

The time complexity of this beam search is
\(
\mathcal{O}\bigl(k^3 + (n-3)\,m\,k\bigr),
\)
where $k^3$ arises from the initial exhaustive search of the first three shots, and $(n-3)\,m\,k$ covers evaluating $k$ extensions for each of the $m$ beam elements for the remaining $(n-3)$ positions. This is substantially more efficient than the naive $\mathcal{O}(k^n)$ search when $m \ll k^{\,n-3}$, making it feasible for real-world multi-shot video assembly tasks.

\section{Experiments}

\subsection{Setup}
\noindent\textbf{Datasets.}
We evaluate \name on two datasets to highlight its versatility. (1) VSPD~\cite{yang2023shot}, the only publicly available dataset specifically designed for shot assembly. After removing videos with fewer than three shots, we have $2,516$ training videos and $124$ testing videos. (2) MSV3C, our curated dataset from V3C (detailed in \cref{sec:dataprep}), which we split $90:10$ into $2,565$ training videos and $285$ testing videos. In the MSV3C test set, each ground-truth shot is augmented with five visually similar alternatives (see \cref{fig:MSC3Ctest}), producing $7,650$ candidate shots for evaluating each assembly method.

\noindent\textbf{Implementation Details.} We implement 
our LCA model using PyTorch~\cite{paszke2019pytorch} and train it on two NVIDIA A100 GPUs. For visual feature extraction, we utilize the CLIP ViT-B/32 model~\cite{dosovitskiy2020image, radford2021learning} pretrained on large-scale image-text pairs. The transformer encoder consists of $4$ layers, $8$ attention heads, a hidden dimension of $512$, and a feedforward dimension of $2,048$. We process each shot by uniformly sampling $5$ frames and maintain a maximum sequence length of $50$ shots.
During training, we employ the AdamW optimizer~\cite{loshchilov2017decoupled} with an initial learning rate of $1\times10^{-3}$. The model is trained for $80$ epochs with a batch size of $128$. For the contrastive learning objective, we set the temperature parameter $\tau$ to $0.1$ and set number of replaced shots $k=40\%$ for negative pair. For \name-text, we set the weight of text alignment \( \gamma \) to $0.5$ using grid search.

\noindent\textbf{Baselines.}
We compare \name with two main groups of methods under the distinct constraints of our datasets. (I) For the \textit{VSPD} dataset, we compare against RATV~\cite{yang2023shot} and Transcript-to-Video~\cite{yang2023shot,xiong2022transcript}, both of which rely on matching text segments (transcripts or captions) to candidate clips, assuming relatively precise scene-level textual annotations. (II) For the \textit{MSV3C} dataset—\textit{which does not have captions for training shot-assembly models}—we benchmark against zero-shot text-video retrieval methods pretrained on large-scale text-video data,
including CLIP4Clip~\cite{luo2021clip4clip}, Frozen-in-Time \cite{bain2021frozen}, and CLIP \cite{radford2021learning}. 

\noindent\textbf{Metrics.} 
We adopt top-1 Recall (R@1), Intersection over Union (IoU), and Sequence Matching Score (SMS) as our evaluation metrics. Recall@K indicates whether shots in the target sequence appear among the top-K predicted sequences; IoU measures the overlap between predicted and ground-truth sequences. For SMS, given a query text $t$, let $\{c_1, c_2, \dots, c_n\}$ denote the ground-truth shot sequence and $\{\tilde{c}_1, \tilde{c}_2, \dots, \tilde{c}_{\tilde{n}}\}$ the predicted sequence. SMS is defined as:
\begin{equation}
\text{SMS} = \frac{1}{n}\sum_{i=1}^{\min(n,\tilde{n})} \mathbb{I}(c_i = \tilde{c}_i)
\end{equation}
where $\mathbb{I}(\cdot)$ is the indicator function, returning 1 if two shots are identical and 0 otherwise~\cite{yang2023shot}.

\subsection{Quantitative Results}
\noindent\textbf{Results on VSPD Dataset.}
Because VSPD lacks direct human ratings, we generate pseudo-labels for training and evaluation by assigning each video a rating from 1 to 5 based on the proportion of substituted clips (starting at 1 for 20\% substitution and incrementing by 1 per additional 20\%). VSPD also provides only coarse, high-level captions—often just a single sentence covering multiple shots—so we adapt our \name pipeline by retrieving the top-15 visually similar shots per caption and initializing our beam search with the top-3 among them.

\Cref{tab:VSPD} shows that both \name and \name-text outperform RATV~\cite{yang2023shot} and Transcript-to-Video~\cite{yang2023shot,xiong2022transcript} on the VSPD dataset, achieving higher video-quality metrics (IoU and SMS) alongside faster inference. Two factors drive these gains. First, since VSPD’s transcripts are not fine-grained scene descriptions, text-driven approaches struggle with mismatched or underspecified textual segments. By contrast, \name primarily enforces shot-to-shot coherence, making it more robust to imprecise text. Second, unlike strictly sequential retrieval used in those baselines, \name holistically optimizes the arrangement of candidate shots, ensuring smoother transitions and better thematic consistency. Notably, \name also cuts average inference time by \textbf{43\%} relative to RATV, underscoring the computational advantages of our beam-search framework.



\noindent\textbf{Results on MSV3C Test Set.}  
Although we manually caption each shot in the MSV3C test set to facilitate evaluation of zero-shot text-video retrieval methods~\cite{luo2021clip4clip,radford2021learning,bain2021frozen}, we do not have corresponding shot-level captions for the training split. Therefore, training text-based assembly models like RATV~\cite{yang2023shot} would require a substantially larger labeled corpus than what is currently available, and thus falls beyond our scope. Instead, we compare \name against zero-shot baselines, which do not require captioned training data, and a random assembly baseline (referred to as Random), which constructs the final video by randomly selecting shots from the candidate pool. \Cref{tab:MSV3C} shows that these text-video retrieval methods perform better here than on VSPD, likely due to the higher quality of our manually generated test captions. Nevertheless, once \name integrates these textual cues, it decisively outperforms all baselines—improving IoU by \textbf{22\%} and R@1 by \textbf{16\%} relative to CLIP4Clip~\cite{luo2021clip4clip}. This result underscores the benefit of explicitly modeling multi-shot coherence when assembling videos, even when detailed text annotations are limited.

\begin{figure*}[h]
    \centering
    \includegraphics[width=0.9\linewidth]{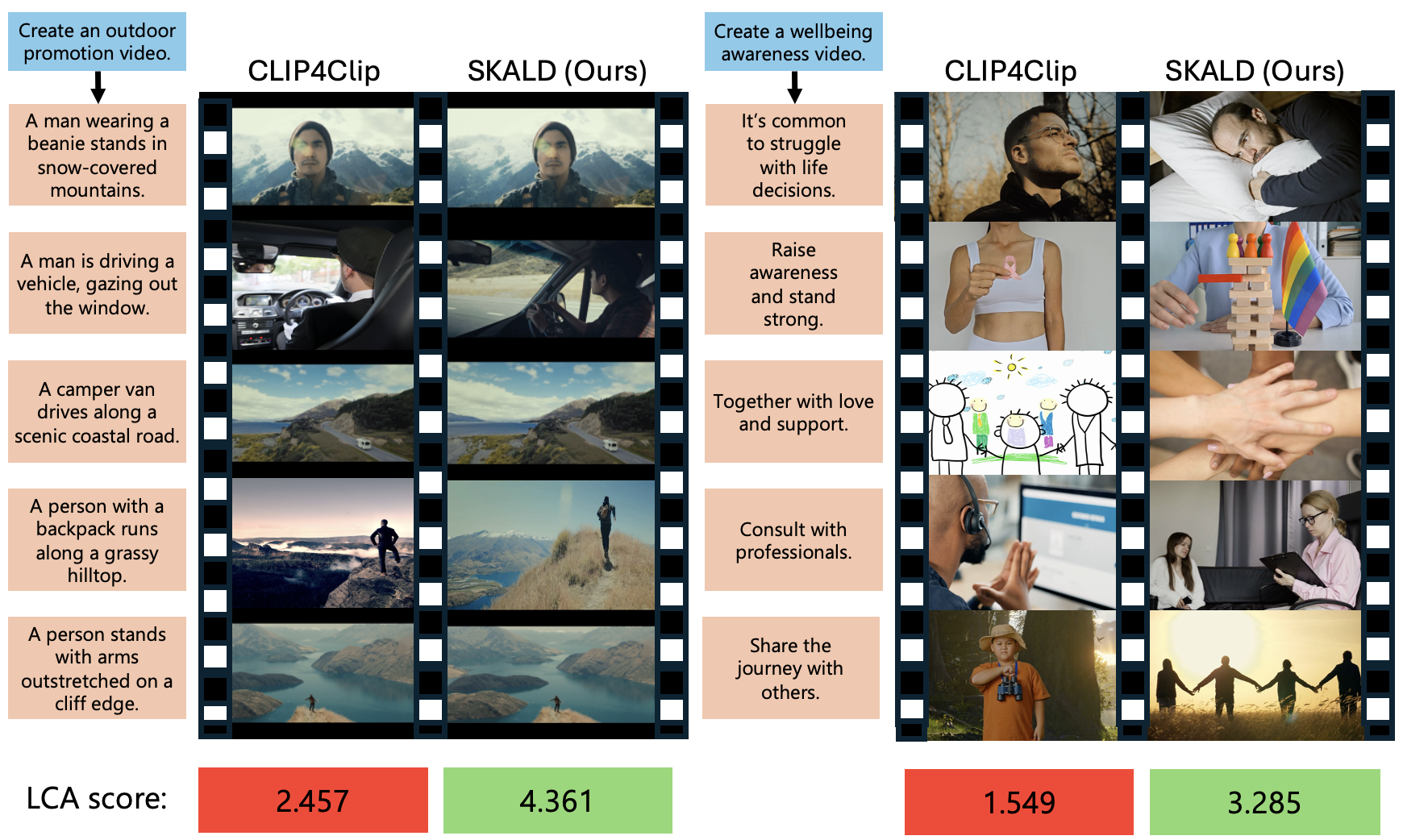}
    \caption{Qualitative comparison of multi-shot assembly.
    Text-only assembly methods (\textit{e.g.}, CLIP4Clip) retrieve shots text-matching shots but disrupt scene continuity with inconsistent landscapes and lighting. In contrast, \name preserves visual and thematic coherence. The LCA scores (green = high, red = low) reflect these differences.}
    \label{fig:qualitative}
\end{figure*}

\begin{table}[t]
   \centering
   \small
   \caption{Quantitative results on the VSPD dataset show that \name outperforms previous coherence-aware video retrieval methods while requiring 43\% less inference time. Incorporating text information further enhances performance.}
   \begin{tabular}{l|cc|c}
   \toprule
   \multirow{2}{*}{Method} & \multicolumn{2}{c|}{Video Quality} & Inference Time \\
   & IoU$\uparrow$ & SMS$\uparrow$ & (s/video)$\downarrow$ \\
   \midrule
        CLIP ViT B/32~\cite{radford2021learning} & 0.072 & 0.034 & 2.153\\
       Frozen-in-Time~\cite{bain2021frozen} & 0.085 & 0.066 & 2.554\\
       Transcript-to-Video~\cite{xiong2022transcript} & 0.096 & 0.064 &  5.152\\ 
       RATV~\cite{yang2023shot} & 0.144 & 0.090 &  6.671\\
       \midrule
       \textbf{\name} & \textbf{0.158} & \textbf{0.114} & 3.745  \\
       \textbf{\name-text} & \textbf{0.214} & \textbf{0.167} & 4.671  \\
   \bottomrule
   \end{tabular}
   \label{tab:VSPD}
\end{table}

\begin{table}[t]
    \small
    \centering
    \caption{Results on the MSV3C test set demonstrate that baseline methods perform significantly better than on VSPD, owing to MSV3C's high-quality shot-level captions. Nonetheless, \name with text incorporation achieves superior performance compared to text-only baselines.}
    \begin{tabular}{l|cc|c}
        \toprule
        \multirow{2}{*}{Method} & \multicolumn{2}{c|}{Video Quality} & Inference Time \\
        & IoU$\uparrow$ & R@1$\uparrow$ & (s/video)$\downarrow$ \\
        \midrule
        Random & 0.102 & 0.164 & 0.015  \\
        CLIP ViT B/32~\cite{radford2021learning}  & 0.205 & 0.341 & 1.675  \\
        Frozen-in-Time~\cite{bain2021frozen} & 0.221 & 0.363 & 1.986\\
        Clip4Clip~\cite{luo2021clip4clip} & 0.269 & 0.424 &  1.622\\
        \midrule
        \textbf{\name} & 0.117 & 0.204 & 4.971 \\
        \textbf{\name-text} & \textbf{0.328} & \textbf{0.494} & 5.779 \\
        \bottomrule
    \end{tabular}
    \label{tab:MSV3C}
\end{table}

\subsection{Qualitative Results}
We present qualitative comparisons between videos assembled by
CLIP4Clip~\cite{luo2021clip4clip} and \name in 
\cref{fig:qualitative},
with shots retrieved from identical queries (shown in the left column). 
CLIP4Clip retrieves shots that match text queries but fails to maintain coherence.
In the left example, it selects a second shot that shifts to a mismatched setting and a fourth shot with inconsistent time of day, disrupting scene continuity. 
In contrast, \name 
preserves visual and narrative flow by selecting contextually aligned shots.
In the right example, CLIP4Clip introduces an animated clip in the third shot, breaking stylistic consistency, and the fourth shot is thematically mismatched (tech vs.~well-being consulting). Meanwhile, \name retrieves in-context 
shots that maintain a cohesive storyline.
The computed LCA scores for these videos confirm these qualitative observations---more coherent videos yield significantly higher scores, validating our metric's effectiveness.

\begin{table}[t]
\small
   \centering
   \caption{Ablation study results demonstrate a clear trade-off between video quality and inference time as beam size increases. Additionally, \name trained with our two-task approach, including the shot coherence task, significantly outperforms end-to-end training without it, even with limited human rating labels.}
   \begin{tabular}{l|cc|c}
   \toprule
   \multirow{2}{*}{Model} & \multicolumn{2}{c|}{Video Quality} & Inference Time \\
   & IoU$\uparrow$ & R@1$\uparrow$ & (s/query)$\downarrow$ \\
   \midrule
   \multicolumn{4}{l}{\textbf{\textit{Beam Size}}} \\
   \name (beam=5)  & 0.114 & 0.191 & 1.602 \\
   \name (beam=20) & 0.117 & 0.204 & 4.971 \\
   \name (beam=50) & 0.123 & 0.218 & 9.254 \\
   \midrule
   \multicolumn{4}{l}{\textbf{\textit{Training Strategy}}} \\
   \name end-to-end & 0.086 & 0.121 & 4.971 \\
   \name two-stage & \textbf{0.117} & \textbf{0.204} & 4.971 \\
   \bottomrule
   \end{tabular}
   \label{tab:ablation}
\end{table}

\begin{table}[t]
    \centering
    \small
    \caption{Ablation study on replaced shot percentage ($k\%$) and length regularization $\lambda$.}
    \resizebox{0.95\columnwidth}{!}{%
    \begin{tabular}{l|ccc|ccc}
    \toprule
    &\multicolumn{3}{c|}{Replaced shots $k\%$} & \multicolumn{3}{c}{Length regularization $\lambda$} \\
    &20 & 40 & 60 & 0.0 & 0.5 & 1.0\\
    \midrule
    IoU & 0.103 & \textbf{0.117} & 0.089  & 0.032 & \textbf{0.117} & 0.108 \\
    \bottomrule
    \end{tabular}
    }
    \label{tab:hyper}
\end{table}

\subsection{Ablation Study}

We conduct ablation studies on models trained with the MSV3C dataset to assess key design choices in \name.
\noindent\textbf{Beam size.} Results in \cref{tab:ablation} 
show that increasing the beam size from 5 to 50 improves both IoU and R@1 metrics, though at the cost of longer inference times, 
highlighting a trade-off between performance and computational cost. 
\noindent\textbf{Two-stage vs. End-to-end.} Our two-stage learning strategy significantly outperforms end-to-end training(0.117 vs. 0.086 IoU, 0.204 vs. 0.121 R@1). Notably, this improvement is achieved with only around 500 human-rated samples, showing that a robust assessment model can be trained without large-scale human ratings.

\noindent\textbf{Replaced shots and length regularization.} We present the impact of replaced shot $k\%$ and length regularization hyperparameter $\lambda$ in \cref{tab:hyper}. 
Replacing 40\% of shots yields the best performance, 
as 20\% 
provides insufficient contrast for negative samples, while 60\% disrupts training.
For $\lambda$, a grid search determined 0.5 as optimal.
Omitting length regularization ($\lambda$ = 0) significantly degrades performance,
highlighting the importance of making the LCA score invariant to sequence length. 


\vspace{-0.1in}
\subsection{User Study}
We conduct a comprehensive user study to compare the video assembly quality of \name and CLIP4Clip~\cite{luo2021clip4clip} wihout audio cues. 
We create 10 video pairs from our stock footage pool, ensuring each pair shares the same title and retrieves shots using identical queries (\cref{fig:assembly_problem}).
In a side-by-side comparison format, 20 participants 
independently rate both videos on a five-point scale, from “Very Poor” to “Excellent,” and we allowed up to 30 minutes for completion (though most finished more quickly). Each pair appears in three separate rating sessions to ensure reliability, preventing individual biases.
Results show a strong preference for \name: \textbf{45\%} of participants preferred videos assembled by \name, whereas only 22\% preferred those created using CLIP4Clip, and the remaining 33\% found both comparable. These outcomes highlight \name's 
superior shot coherence and visual storytelling over text-based retrieval. Additional details are in the supplementary material.



\vspace{-0.1in}
\section{Conclusion}
\vspace{-0.1in}
We present \name, a multi-shot video assembly framework designed to create coherent video sequences from candidate shots with minimal reliance on video captions, while optionally incorporating textual cues through its \name-text variant. At the core of \name is the \emph{Learned Clip Assembly} (LCA) score, a metric trained to quantify narrative coherence by modeling temporal and semantic relationships between shots. Our contrastive pretraining for shot coherence followed by supervised regression enables efficient training even under limited annotation budgets. Leveraging beam search enhanced with sequence-length decorrelation, \name efficiently navigates the combinatorial shot space without exhaustive enumeration. Experiments on the MSV3C and VSPD datasets demonstrate substantial improvements over prior methods in both video quality and computational efficiency, validated by a user study indicating clear subjective preference. 
We aim to apply \name to longer-form contents as future work.

\section*{Acknowledgement}
This material is based in part upon work supported by the National Science Foundation under 
NSF Grant Numbers CNS-2146449 (NSF-CAREER)
and CNS-2333487 (NSF-FRONTIER) and gift funding from Adobe Research. Any opinions, findings, and conclusions or recommendations expressed in this material are those of the authors and do not necessarily reflect the views of the sponsors.

{
    \small
    \bibliographystyle{ieeenat_fullname}
    \bibliography{main}
}

\clearpage
\setcounter{page}{1}

\maketitlesupplementary

\begin{figure*}[h]
    \centering
    \includegraphics[width=0.9\linewidth]{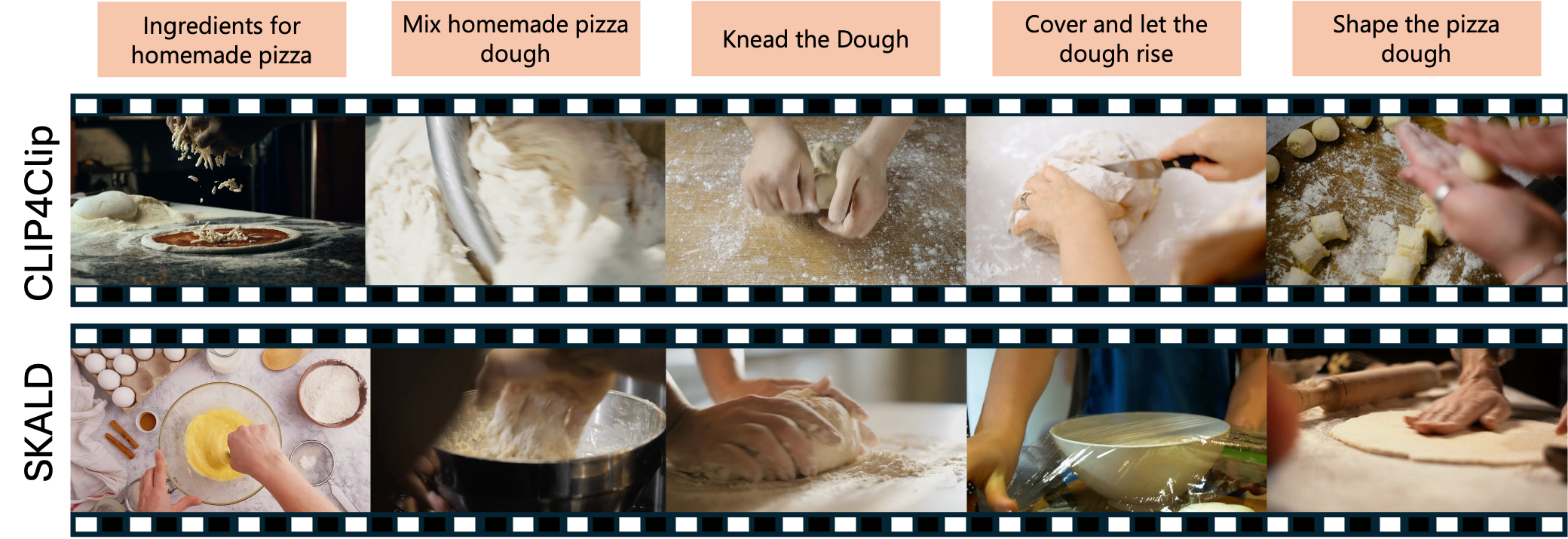}
    \caption{Qualitative analysis shows that standard text-based retrieval methods readily match individual pizza-making steps but struggle to maintain cohesive transitions between them. By contrast, our approach consistently assembles visually coherent sequences that accurately follow each stage of dough preparation. }
    \label{fig:quali_supp_01}
\end{figure*}

\begin{figure*}[h]
    \centering
    \includegraphics[width=0.9\linewidth]{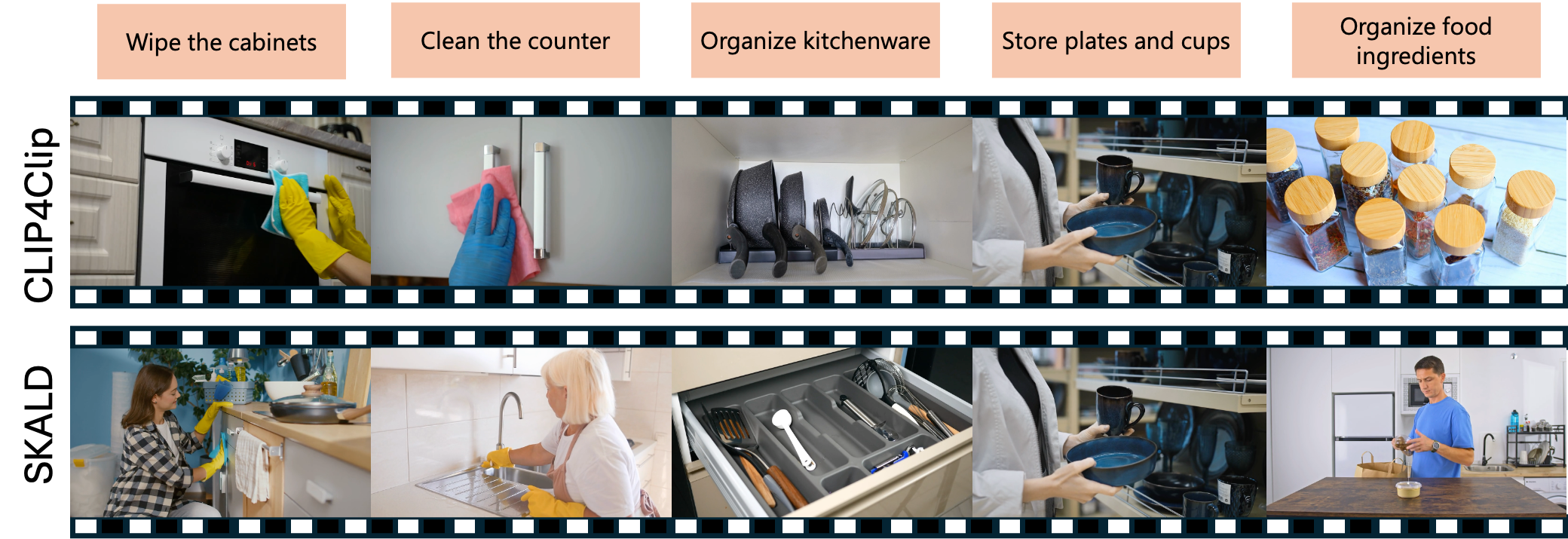}
    \caption{Qualitative analysis indicates that basic text-based methods identify the individual cleaning tasks but often lack seamless transitions between them. In contrast, our approach constructs visually coherent sequences that reflect each stage of kitchen organization. }
    \label{fig:quali_supp_02}
\end{figure*}

\begin{figure*}[h]
    \centering
    \includegraphics[width=0.9\linewidth]{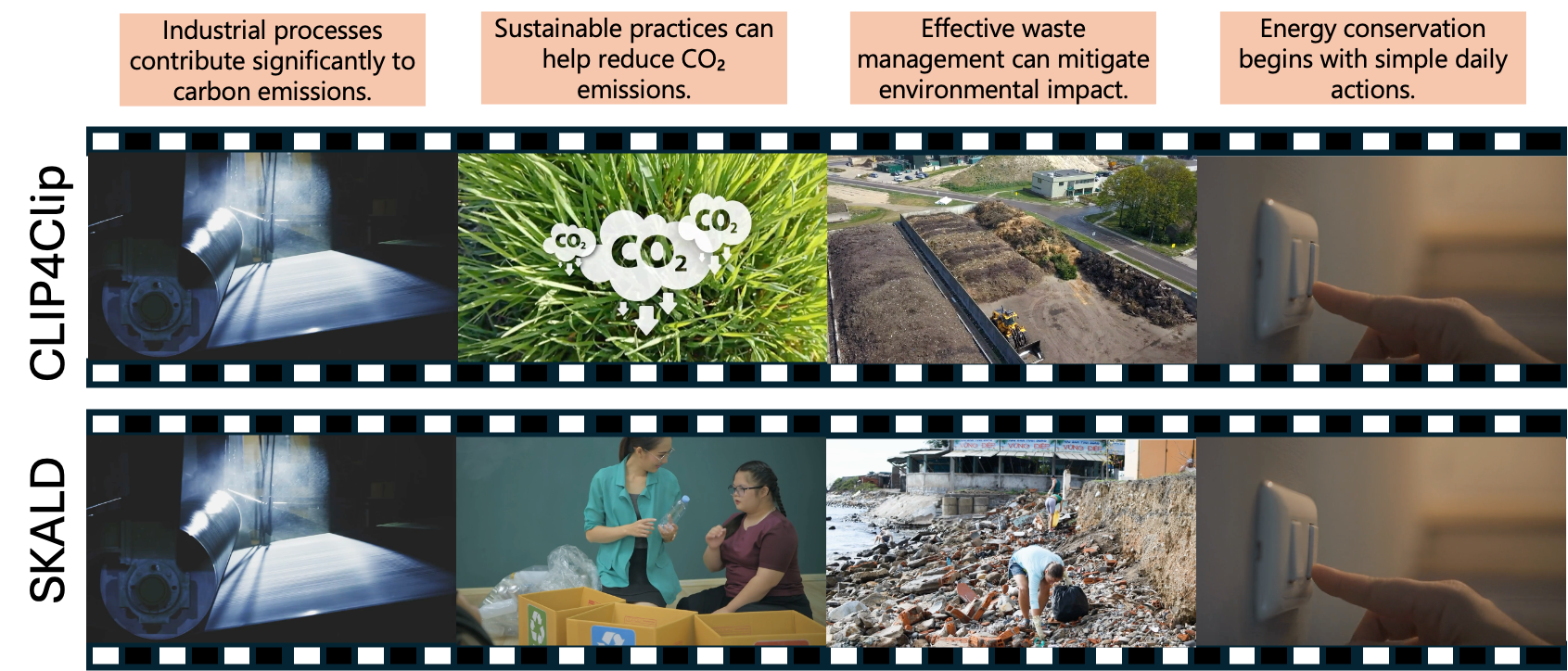}
    \caption{Qualitative analysis indicates that basic text-based methods identify the individual cleaning tasks but often lack seamless transitions between them. Nonetheless, our method ensures smooth transition between shots for a promotion video on environment awareness.}
    \label{fig:quali_supp_03}
\end{figure*}

\section{More Details on Data Preparation}

\paragraph{MSV3C training set.} We collected videos from the V3C1 dataset~\cite{rossetto2019v3c}, comprising $28,450$ videos of $2-5$ minutes duration. While V3C1 provides scene segmentation, we found its scene cuts did not accurately align with shot transitions. Therefore, we used TransNetV2~\cite{souvcek2020transnet} to split videos into individual shots. We filtered out shots shorter than 2 seconds, static frames, and shots containing excessive text overlays. To ensure video diversity, we computed the average CLIP embedding for each shot and analyzed variance across videos. We excluded the lowest variance quartile, typically consisting of interview-style videos with limited visual diversity. For each remaining video, we randomly sampled between 3 and 12 consecutive shots, forming representative multi-shot sequences suitable for training and evaluation. The resulting dataset encompasses 5,648 carefully curated videos, totaling over 12,000 shots.

\paragraph{MSV3C testing set.} We divided the MSV3C dataset into training and testing sets with a 9:1 ratio. To simulate real-world video editing scenarios, each shot in the test set was manually annotated with a concise single-sentence scene description. These descriptions served as textual queries to retrieve five visually similar candidate shots per original shot from our internal stock footage collection. This design closely reflects practical shot assembly workflows, where the objective is identifying the optimal coherent sequence among visually and semantically related alternatives.

\section{User Study}

We conducted a comprehensive user study to rigorously assess and compare the video assembly quality between CLIP4Clip~\cite{luo2021clip4clip} and \name. The primary objective was evaluating these methods' ability to generate coherent and engaging visual narratives without relying on audio elements.

Our study employed a side-by-side comparison format. Participants simultaneously viewed two multi-shot videos sharing identical thematic content. Each video explicitly excluded voice-overs and music, enabling unbiased evaluation of visual storytelling proficiency. Participants independently rated the quality of each video on a standardized five-point scale from ``Very Poor" to ``Excellent," ensuring precise quantification of their qualitative judgments. Participants were provided with ample time—up to 30 minutes—although task completion typically required less than half that duration.  

To maintain data reliability, ``golden set" video pairs with predetermined ratings were included as quality controls. Submissions failing these controls were excluded. The study involved 3 distinct rating sessions covering 10 unique video pairs, with each pair evaluated multiple times by different participants. This robust redundancy mitigated subjective bias, ensuring comprehensive evaluation.

\section{More Qualitative Results}

The qualitative analysis presented in~\cref{fig:quali_supp_01},~\cref{fig:quali_supp_02}, and~\cref{fig:quali_supp_03} underscores that \name consistently outperforms conventional text-based methods in preserving visual continuity and narrative coherence. As illustrated in~\cref{fig:quali_supp_01}, traditional text-based retrieval successfully identifies individual pizza-making steps but frequently introduces visual discontinuities such as inconsistent lighting conditions or contextually irrelevant scenes. Conversely, \name reliably assembles sequences that flow logically through each cooking stage, sustaining thematic unity. 
Similarly, in~\cref{fig:quali_supp_02}, standard retrieval methods often correctly pinpoint discrete kitchen-cleaning tasks yet fail to ensure seamless visual and contextual transitions. \name excels by selecting visually coherent and contextually consistent shots, effectively conveying a natural narrative progression. Finally, the environmental awareness promotion in~\cref{fig:quali_supp_03} exemplifies \name's capacity to preserve smooth transitions across visually distinct but semantically connected scenes, significantly enhancing viewer engagement through sustained narrative coherence.

\section{Video Demo}

We provide three example videos (\texttt{0001.mp4, 0002.mp4, 0003.mp4}) assembled with \name. The detailed themes and descriptions corresponding to each scene within these videos are comprehensively documented in the accompanying file \texttt{video demo.rtf}. These examples demonstrate \name’s practical capability in generating high-quality, thematically cohesive multi-shot videos across diverse scenarios.

\end{document}